%% file: main.tex
\documentclass[10pt,twocolumn,letterpaper]{article}
\usepackage{float}             
\usepackage{placeins}
\usepackage{multirow}
\usepackage[T1]{fontenc}

\usepackage{fancyhdr}

\usepackage[pagenumbers]{cvpr} 

\input{preamble}

\definecolor{cvprblue}{rgb}{0.21,0.49,0.74}
\usepackage[pagebackref,breaklinks,colorlinks,citecolor=cvprblue]{hyperref}

\def\paperID{4} 
\def\confName{CVPR}
\def\confYear{2024}

\title{Feature Corrective Transfer Learning: End-to-End Solutions to Object Detection in Non-Ideal Visual Conditions}

\author{Chuheng Wei,\qquad   Guoyuan Wu,\qquad   Matthew J. Barth\\
University of California Riverside\\
{\tt\small chuheng.wei@email.ucr.edu}
}

\begin{document}

\makeatletter
\g@addto@macro\@maketitle{
  \begin{figure}[H]
  \setlength{\linewidth}{\textwidth}
  \setlength{\hsize}{\textwidth}
  \centering
    \includegraphics[width=1\linewidth]{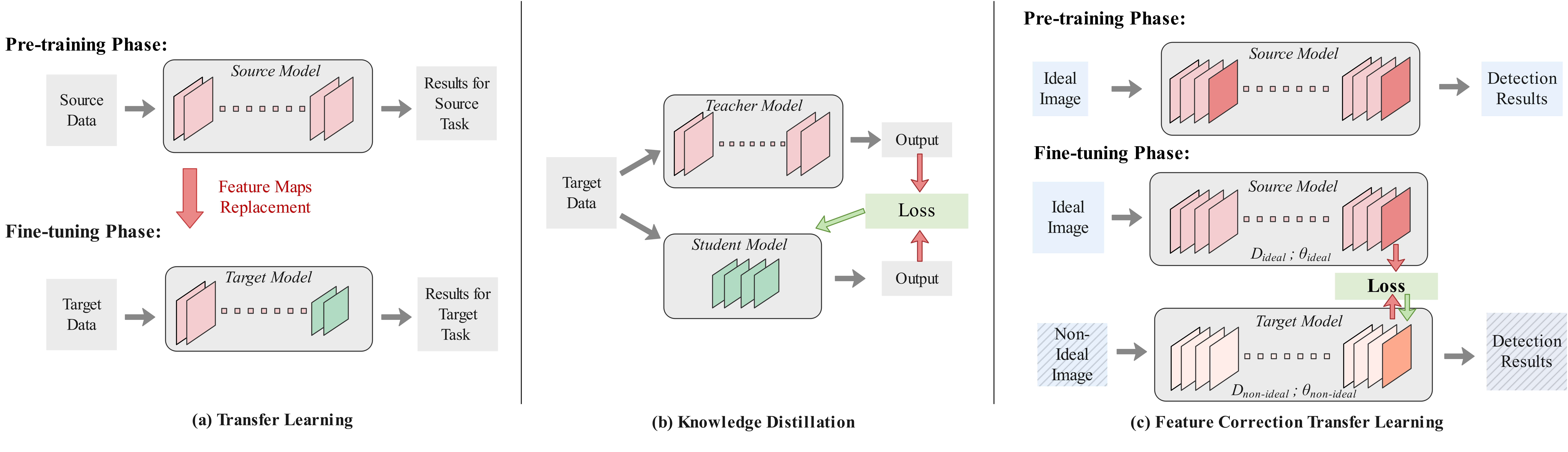}
    \caption{A Visual Comparison of (a) Transfer Learning, (b) Knowledge Distillation, and (c) Feature Correction Transfer Learning}
    \label{fig:FCTL}
  \end{figure}
}

\makeatother

\maketitle

\input{sec/0_abstract}

\input{sec/1_intro}
\input{sec/2_relatedwork}

\input{sec/3_method}
\input{sec/4_experiments}

\input{sec/5_Conclusion}

{
    \small
    \bibliographystyle{ieeenat_fullname}
    \bibliography{main}
}


\end{document}

%% file: preamble.tex
\usepackage[dvipsnames]{xcolor}


%% file: sec/0_abstract.tex
\begin{abstract}

A significant challenge in the field of object detection lies in the system's performance under non-ideal imaging conditions, such as rain, fog, low illumination, or raw Bayer images that lack ISP processing. Our study introduces ‘Feature Corrective Transfer Learning’, a novel approach that leverages transfer learning and a bespoke loss function to facilitate the end-to-end detection of objects in these challenging scenarios without the need to convert non-ideal images into their RGB counterparts.
In our methodology, we initially train a comprehensive model on a pristine RGB image dataset. Subsequently, non-ideal images are processed by comparing their feature maps against those from the initial ideal RGB model. This comparison employs the Extended Area Novel Structural Discrepancy Loss (EANSDL), a novel loss function designed to quantify similarities and integrate them into the detection loss. This approach refines the model’s ability to perform object detection across varying conditions through direct feature map correction, encapsulating the essence of Feature Corrective Transfer Learning.
Experimental validation on variants of the KITTI dataset demonstrates a significant improvement in mean Average Precision (mAP), resulting in a 3.8-8.1\% relative enhancement in detection under non-ideal conditions compared to the baseline model, and a less marginal performance difference within 1.3\% of the mAP@[0.5:0.95] achieved under ideal conditions by the standard Faster RCNN algorithm.

\end{abstract}

%% file: sec/1_intro.tex
\section{Introduction}
\label{sec:intro}


As a vital component of computer vision, object detection is used in a wide range of applications such as autonomous driving, surveillance, and augmented reality \cite{zou2023object}. 
Despite significant advancements, the robust detection of objects under non-ideal imaging conditions—such as rain \cite{Sindagi2019Prior-Based}, fog \cite{wang2019saliencygan}, low illumination \cite{cui2021multitask}, or directly from raw Bayer images \cite{bayer1976color} without Image Signal Processing (ISP) \cite{wei2022vehicle}—remains a considerable challenge. 
Traditional methods often rely on preprocessing steps to convert non-ideal images into more ‘ideal’ conditions before detection \cite{liu2022image}, which can lead to loss of details and introduce unwanted artifacts. 


In response to these challenges, transfer learning presents an effective strategy by leveraging pre-existing models trained on extensive, well-labeled datasets to address the variances in imaging conditions \cite{zhuang2020comprehensive}.
 Traditionally, it involves a two-phase process: initially training a source model on comprehensive source data, followed by fine-tuning where parts of this model are adjusted or fixed to adapt to new tasks as shown in Figure \ref{fig:FCTL}-(a). While effective in managing image quality variances, existing transfer learning approaches often fall short in addressing the specific challenges posed by non-ideal imaging environments \cite{talukdar2018transfer}.

Knowledge distillation algorithms \cite{hinton2015distilling}, as shown in Figure \ref{fig:FCTL}-(b), primarily utilize a complex model (Teacher) pretrained on a large dataset  to enhance the performance of a smaller model (Student) on a target task. This is achieved by minimizing the loss function based on the discrepancy between their outputs, thus guiding the student model's improvement. Drawing inspiration from both traditional transfer learning and knowledge distillation, we propose the \textit{Feature Corrective Transfer Learning} (FCTL) approach, illustrated in Figure \ref{fig:FCTL}-(c). In the pre-trained phase, a comprehensive source model is trained on ideal images. During the fine-tuning phase, the structure and parameters of the source model are kept unchanged while establishing an identical target model. This phase involves training with both non-ideal and ideal versions of the same image, leveraging the established source model to perform feature correction on the target model through a specific loss function at selected layers. This novel approach, FCTL, distinguishes itself by emphasizing direct feature map correction to enhance the robustness and accuracy of object detection models under non-ideal conditions, without necessitating the conversion of non-ideal images to their RGB counterparts.

Based on this framework, we have developed the \textit{Feature Corrective Transfer Learning} NITF-RCNN algorithm that is supplemented by the Extended Region Novel Structure Difference Loss (EANSDL). In our method, we use a two-stage training strategy, establishing a strong baseline using the original RGB dataset and then performing feature map correction on non-ideal image models. By prioritizing direct feature map correction over traditional preprocessing, this process iteratively enhances the model's ability to detect objects under adverse conditions.


\subsection{Contributions}

\begin{itemize}
    \item \textbf{Feature Corrective Transfer Learning Framework:} A new transfer learning approach is tailored to object detection in challenging conditions, employing feature map correction to align non-ideal images with high-quality RGB datasets, thereby improving robustness and detection accuracy.

    \item \textbf{Non-Ideal Image Transfer Faster-RCNN (NITF-RCNN):} An adaptation of the Faster-RCNN architecture incorporates our feature map correction algorithm, designed to specifically address the challenges presented by non-ideal imaging conditions, ensuring a thoughtful rather than blanket application of transfer learning.

    \item \textbf{Extended Area Novel Structural Discrepancy Loss (EANSDL):} A novel loss function is created to facilitate feature map correction, enabling precise adjustments during training by quantifying the discrepancy between feature maps under different conditions, thus enhancing the model's performance in detecting objects across diverse visual environments.
\end{itemize}



%% file: sec/2_relatedwork.tex
\section{Related Work}
\label{sec:relatedwork}
\subsection{Object Detection under Non-Ideal Visual Conditions}


Within the domain of object detection under non-ideal visual conditions, a diverse array of strategies has been explored, ranging from traditional preprocessing to innovative end-to-end models. Sindagi et al. \cite{Sindagi2019Prior-Based} preprocessed images affected by haze and rain using traditional techniques and weather-specific knowledge for object detection. Kvyetnyy et al. \cite{Kvyetnyy2017Object}, alternatively, addressed low-light challenges through denoising methods like bilateral filtering and wavelet thresholding, aiming to improve detection performance. Moreover, some scholars have adopted two-stage model approaches, grounded in deep learning. For example,
Huang et al. \cite{huang2020dsnet} introduced a dual-subnet network (DSNet), comprising detection and restoration subnets to achieve image restoration and object detection under harsh weather conditions separately. Yang et al. \cite{Yang2023A} presented a two-stage unsupervised deraining approach, utilizing non-local contrastive learning to decouple the rain layer from clean images more effectively before object detection tasks.

Emerging research, however, aims to develop truly end-to-end solutions. For example, Wang et al. \cite{wang2022end} proposed an end-to-end object detection network to mitigate the impact of rainfall, featuring cascaded networks for image restoration and object detection. While this is an end-to-end training approach, deraining and detection are still separated into two networks, with the first network producing derained images.  Additionally, Wei et al. \cite{wei2023enhanced} attempted to perform end-to-end object detection on raw images by incorporating camera parameters into the network to adapt to the features of raw Bayer images. However, This method requires other forms of input besides images and more complex neural networks.

In summary, there is a conspicuous absence of suitable, truly end-to-end models that forego the intermediate image restoration step. There also lacks a universal model capable of effectively handling all types of non-ideal conditions, highlighting a significant gap in the current state of object detection under non-ideal visual conditions.

\subsection{Advancing Object Detection through Transfer Learning Techniques}

In the field of computer vision, transfer learning has emerged as a key strategic tool for improving performance in related, yet distinct, tasks \cite{talukdar2018transfer}. To enhance model performance and efficiency, researchers have begun applying the principles of transfer learning to object detection since its introduction. To improve the accuracy of object detection, Ito et al. \cite{Ito2022Transfer} used genetic algorithms within the transfer learning process to determine which layers should be re-learned automatically. They also avoided the trial-and-error approach inherent in traditional approaches to object detection. Their research demonstrates that partially intercepting neural networks can enhance the efficiency of transfer learning

Several studies have focused on transferring abilities learned from large, general datasets to more specialized tasks, such as detecting small objects. A resolution adaptation scheme was employed by Xu et al. \cite{Xu2023TranSDet:} to enhance the detection of small-scale objects by adjusting models trained on generic datasets using images of various smaller resolutions, thereby significantly increasing performance. According to Bu et al. \cite{Bu2021GAIA:}, a transfer learning system named \textit{GAIA} was designed in recognition of the unique requirements of the object detection domain. Using this system, tailored solutions are automatically generated based on heterogeneous downstream demands, offering powerful pretrained weights and selecting models that meet specific needs for object detection, including latency constraints and particular data domains, thereby demonstrating a significant advancement in small-sample object detection algorithms.

Additionally, efforts have been made to extend the utility of transfer learning beyond mere knowledge transfer by utilizing data augmentation and synthetic datasets. Talukdar et al. \cite{talukdar2018transfer} explored the generation of synthetic datasets through various data augmentation algorithms to assist in the transfer learning process for convolutional neural networks. Their experiments across a range of object detection algorithms validated the significance of synthetic datasets in enhancing transfer learning outcomes. Moreover, some researchers have applied transfer learning to address challenges presented by non-ideal data conditions. For instance, Chen et al. \cite{Chen2021Object} improved the Faster R-CNN algorithm through transfer learning, employing two domain adaptation structures to measure domain similarity. Their \textit{Domain Adaptation Faster R-CNN} which utilized adversarial training, proved effective in low-light conditions.

However, these approaches primarily focus on domain-specific challenges and do not extensively explore object detection under adverse weather conditions through transfer learning. This gap highlights the novelty of the proposed framework, which specifically addresses feature map correction for object detection in challenging environmental conditions, setting a new research direction in this area.

%% file: sec/3_method.tex
\begin{figure*}[ht]
    \centering
    \includegraphics[width=1\linewidth]{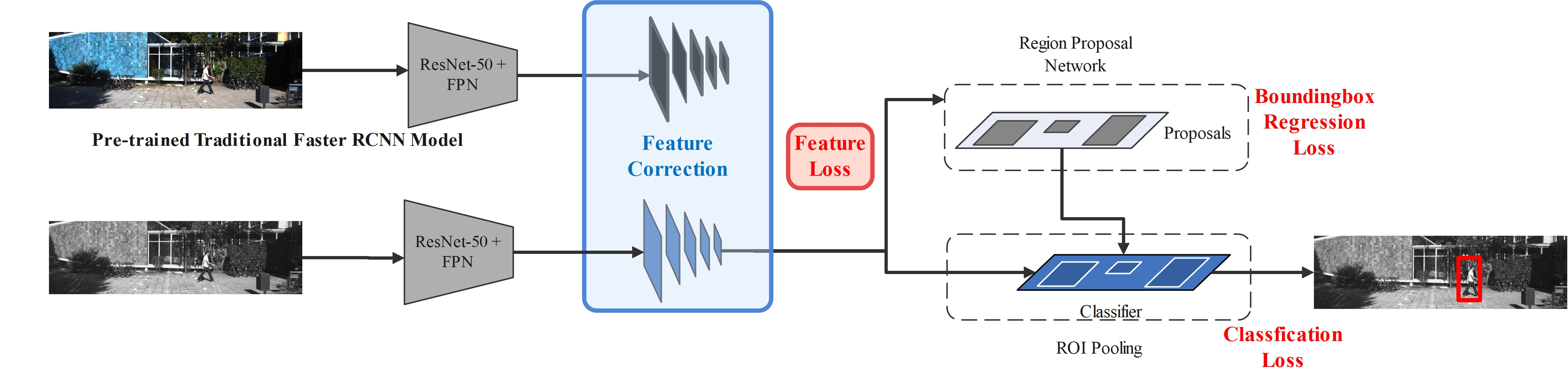}
    \caption{ Architecture of Non-Ideal Image Transfer Faster RCNN (NITF-RCNN) Model}
    \label{fig:NITF-RCNN}
\end{figure*}

\section{Feature Corrective Transfer Learning Framework}
\label{sec:feature_correction_framework}

End-to-end object detection holds paramount importance in computer vision, offering a seamless process from image input to object identification and localization \cite{carion2020end}. Traditionally, object detection under non-ideal conditions such as poor lighting, adverse weather, or unprocessed raw images, necessitates transforming these images into a more 'ideal' state before detection can occur. However, this transformation often targets human visual preferences rather than the requirements of neural networks. This paper posits that true end-to-end detection should circumvent the need for such transformations, thereby directly addressing the detection in non-ideal conditions.

However, evidence suggests that models developed for ideal situations do not perform optimally on non-ideal cases \cite{chan2023raw}, which underscores the necessity for model adjustments \cite{wei2023enhanced}. Direct modifications to handle non-ideal image features could introduce bias or overfitting to specific conditions. To address this, we propose \textit{Feature Map Correction} (FMC) to assist the neural network training process without altering the underlying architecture. Most object detection algorithms process the input image through a neural network to perform bounding box regression and object classification across various feature layers and scales. The \textit{Feature Corrective Transfer Learning} (FCTL) method introduced in this paper aims to guide the training of models on non-ideal images towards closer alignment with the feature layers of models trained on ideal images.

\subsection{Implementation of the FCTL Framework}

To formalize the FCTL framework, we define a mathematical model consisting of the following key steps:

\begin{enumerate}
    \item \textbf{Model Selection and Training on Ideal Images}
    
    Let \(D_{\text{ideal}}\) represent the dataset of ideal images. We select an object detection model \(M\), and train it on \(D_{\text{ideal}}\) to optimize the model parameters \(\theta_{\text{ideal}}\):
    \begin{align}
     \theta_{\text{ideal}} = \arg\min_\theta \mathcal{L}_{\text{det}}(M(D_{\text{ideal}}; \theta)) 
     \end{align}
    where \(\mathcal{L}_{\text{det}}\) is the total loss function for the object detection task, typically comprising classification loss and bounding box regression loss.
    
    \item \textbf{Generation of Non-Ideal Image Versions}
    
    For each ideal image \(x \in D_{\text{ideal}}\), we generate a non-ideal version \(x'\) by synthesizing non-ideal conditions such as rainy weather. This can be achieved by adding noises, blurring, etc.
    
    \item \textbf{Training the Same Object Detection Model on Non-Ideal Images}
    
    Next, the same model \(M\) is trained on the non-ideal images \(D_{\text{non\_ideal}}\), while also using the corresponding ideal images for validation. During this phase, one or multiple feature layers are selected to assess the similarity between the feature maps of the model trained on ideal images and those of the model being trained on non-ideal images, employing a feature similarity loss function \(\mathcal{L}_{\text{fs}}\):

    \begin{align}
     \theta_{\text{non\_ideal}} = \arg\min_\theta \bigg( 
     & \mathcal{L}_{\text{det}}(M(D_{\text{non\_ideal}}; \theta))  \nonumber \\
     & + \lambda \mathcal{L}_{\text{fs}}(F_{\text{ideal}}, F_{\text{non\_ideal}}) \bigg).
    \end{align}
    Here, \(F_{\text{ideal}}\) and \(F_{\text{non\_ideal}}\) represent the feature maps from the model under ideal and non-ideal conditions, respectively, and \(\lambda\) is a coefficient that balances the two loss terms.
    
    \item \textbf{Incorporating Feature Similarity Loss during Backpropagation}
    
    During the backpropagation process in training, the total loss \(\mathcal{L}_{\text{total}}\) to be minimized includes not only the standard detection loss \(\mathcal{L}_{\text{det}}\)—comprising classification loss and bounding box regression loss—but also the feature similarity loss \(\mathcal{L}_{\text{fs}}\). This dual-objective loss function aims to ensure accuracy in object detection while introducing a mechanism for feature map correction:
    \begin{align}
    \mathcal{L}_{\text{total}} = \mathcal{L}_{\text{det}} + \lambda \mathcal{L}_{\text{fs}} ,
    \end{align}
    \begin{align}
     \theta = \arg\min_\theta \mathcal{L}_{\text{total}} . 
    \end{align}
\end{enumerate}

The feature similarity loss \(\mathcal{L}_{\text{fs}}\) is designed to effectively measure the discrepancies in structure and content between the feature maps of the model trained on ideal images and those trained on non-ideal images. It is crucial to note that similarity in feature space can significantly differ from image similarity, necessitating a distinct evaluation metric. This paper introduces the \textit{Extended Area Novel Structural Discrepancy Loss} (EANSDL) method to assess the similarity at the feature level.

In subsequent section, we elaborate on the modifications to the Faster RCNN framework, leading to the development of the Non-Ideal Image Transfer Faster-RCNN (NITF-RCNN) model. This model incorporates a feature similarity loss to evaluate the similarity of pyramid feature maps, showcasing the practical application of the FCTL methodology.

\section{Methodology}
\label{sec:methods}

\subsection{Non-Ideal Image Transfer Faster R-CNN (NITF-RCNN)}
\label{subsec:NITFRCNNframework}

The NITF-RCNN framework adapts the traditional Faster R-CNN \cite{girshick2015fast} to object detection in non-ideal visual conditions, maintaining the original architecture while incorporating feature map correction. As shown in Figure \ref{fig:NITF-RCNN}, this specialized algorithm employs a dual backbone structure: a static backbone derived from a model pre-trained on ideal images and a dynamic backbone that is fine-tuned on non-ideal images.

\noindent\textbf{Training on Ideal Images }

The foundation is laid by training a Faster R-CNN model equipped with a ResNet-50 backbone \cite{he2016deep} and Feature Pyramid Network (FPN) \cite{lin2017feature} on a dataset of ideal RGB images to establish the static backbone. The objective function for this training phase is defined as:
\begin{align}
\theta_{\text{ideal}} = \arg\min_\theta \mathcal{L}_{\text{Faster R-CNN}}(D_{\text{ideal}}; \theta).
\end{align}

\noindent\textbf{Feature Corrective Transfer Learning}

In the feature-level transfer learning phase, the pre-trained static backbone extracts feature maps from the ideal images to form the "ideal pyramid," while the non-ideal images are concurrently processed through the dynamic backbone. Both are then subject to the region proposal network (RPN) \cite{girshick2015fast} and Region of Interest (ROI) Pooling \cite{girshick2014rich}. The dynamic backbone undergoes training, guided by the following combined loss function:
\begin{align} 
\mathcal{L}_{\text{total}} = \mathcal{L}_{\text{det}}(D_{\text{non\_ideal}}; \theta) + \lambda \mathcal{L}_{\text{EANSDL}}(F_{\text{ideal}}, F_{\text{non\_ideal}}).
\end{align}
where $\mathcal{L}_{\text{det}}$ represents the standard object detection loss, which includes bounding box regression and classification losses \cite{girshick2015fast}. $\mathcal{L}_{\text{EANSDL}}$ denotes the \textit{Extended Area Novel Structural Discrepancy Loss}, a dedicated loss function assessing the similarity between feature maps. $\lambda$ serves as the balancing coefficient for the feature similarity loss.

\noindent\textbf{Validation}

During validation, only the non-ideal images are processed through the NITF-RCNN to assess the model's detection capability in non-ideal conditions. This is achieved by applying the trained model to non-ideal images, with the loss functions serving as indicators of performance:
\begin{align} 
\mathcal{L}_{\text{validation}}(D_{\text{non\_ideal}}; \theta).
\end{align}
\noindent\textbf{Structural Summary and Potential Advantages}

The NITF-RCNN framework retains the integrity of the traditional Faster R-CNN structure, with the addition of the feature correction component for ideal images. This approach provides several potential advantages:
\begin{itemize}
    \item \textbf{Feature Correction Without Structural Modification:} The framework enhances object detection under non-ideal conditions without the need for significant modifications to the existing architecture.
    \item \textbf{Direct Feature-Level Adaptation:} By correcting feature maps directly through the FCTL approach, the model is better equipped to handle environmental disturbances inherent in non-ideal images.
    \item \textbf{Balanced Learning:} The use of a joint loss function allows the model to balance feature correction with the primary detection tasks, potentially improving generalization and robustness.
\end{itemize}

\subsection{Adaptive Structural Alignment via EANSDL}

To address the challenge of aligning and comparing feature maps under complex, non-ideal conditions, we introduce \textit{Extended Area Novel Structural Discrepancy Loss} (EANSDL) which not only identifies pixel-level discrepancies but also ensures the structural integrity across larger areas, making it beneficial for advanced object detection frameworks like Faster RCNN in less-than-ideal conditions. EANSDL conducts a comprehensive evaluation, rectifying immediate discrepancies while maintaining overall structural coherence, significantly enhancing Faster RCNN's detection precision. Its design adaptively balances the analysis between detailed discrepancies and broader alignments, dynamically adjusting the gradient consistency evaluation across the feature pyramid's hierarchical layers. This method achieves heightened sensitivity to layer-specific scales and resolutions, thereby bolstering structural integrity and ensuring robust object detection across diverse imaging conditions.

\subsubsection{Mathematical Formulation}
Consider two feature maps, $A$ (non-ideal conditions) and $B$ (ideal conditions), each with dimensions $[batch size, channels, width, height]$. The formulation of EANSDL is given by:

\begin{align}
&\text{EANSDL}(A, B, \delta, r_\mathcal{L})\nonumber \\
&= D(\delta) \cdot \frac{1}{W \cdot H} \sum_{x=1}^{W} \sum_{y=1}^{H} \bigg( \exp(-\Delta S(x, y)) \cdot \Delta S(x, y) \nonumber \\
& + \lambda \cdot \Omega(A, B, x, y, r_\mathcal{L}) \bigg).
\end{align}

where:
\begin{itemize}
\item $D(\delta)$ introduces a time-varying attenuation factor that adjusts the sensitivity of the loss function to training progress,with $\tau$ denoting the ratio of the current epoch to total epochs (ensuring that the loss adapts throughout the training lifecycle).
\item $\Delta S(x, y)$ denotes the local gradient magnitude difference at position $(x, y)$, capturing immediate structural variances.
\item $\lambda$ represents a balancing factor for the contribution of the extended area gradient consistency.
\item $\Omega(A, B, x, y, r_\mathcal{L})$ encapsulates the extended area gradient consistency across a neighborhood radius $r_\mathcal{L}$, dynamically adjusted for each level of the Faster RCNN feature pyramid as:
\begin{equation}
r_{\mathcal{L}} = r_0 / 2^{level},
\end{equation}

where $r_0$ is the initial radius at the largest feature map, and $level$ denotes the specific layer within the feature pyramid.
\end{itemize}

\subsubsection{Implementation Details}

\noindent\textbf{Time-varying Attenuation Factor}

The \textit{Time-varying Attenuation Factor}, represented as $D(\delta)$, introduces a dynamic mechanism to adjust the responsiveness of the loss function throughout the training duration. The term $\delta$ signifies the proportion of the current epoch relative to the total number of epochs, calculated as:

\begin{equation}
\delta = \frac{\text{current\_epoch}}{\text{total\_epochs}}. 
\end{equation}

The implementation of this factor facilitates a methodological shift in the model's emphasis from rectifying prominent structural disparities in the initial training phase to honing finer details in subsequent phases of the training process.

$D(\delta)$ is delineated as follows:

\begin{align}
D(\delta) = \exp(-\alpha \cdot \delta^\beta),
\end{align}

where:
\begin{itemize}
\item $\alpha$ regulates the initial steepness of the decay trajectory;
\item $\beta$ adjusts the curvature to slow down the decay pace.
\end{itemize}

Fundamentally, $D(\delta)$ empowers the model to initially concentrate on correcting significant mismatches between feature maps, ensuring a solid foundation is established. As training progresses and the model evolves in complexity, the attenuation factor reduces the emphasis on these mismatches.
This modification aids in diminishing the influence of the EANSDL on the total loss for object detection during the later phases, allowing for a greater focus on the quintessential tasks of object detection.


\noindent\textbf{Gradient Computation Function }

Central to EANSDL, the gradient computation function $G(\cdot)$ employs the Sobel operator \cite{sobel1968}  to delineate edges and structural attributes across the feature maps. This operator convolves the feature map with two distinct 3$\times$3 kernels, each engineered to unearth edges along respective orientations:

\begin{equation}
S_x = \begin{bmatrix} -1 & 0 & 1 \\ -2 & 0 & 2 \\ -1 & 0 & 1 \end{bmatrix},
\quad
S_y = \begin{bmatrix} -1 & -2 & -1 \\ 0 & 0 & 0 \\ 1 & 2 & 1 \end{bmatrix}.
\end{equation}

Vertical edges are identified through the horizontal gradient (Sobel-x), whereas horizontal edges are pinpointed by the vertical gradient (Sobel-y). The formulas for calculating these gradients are as follows:

\begin{equation}
G_x(A) = A * S_x, \quad G_y(A) = A * S_y.
\end{equation}

The aggregate gradient magnitude is consequently determined by amalgamating these orthogonal gradients:

\begin{equation}
G(A, x, y) = \sqrt{G_x(A, x, y)^2 + G_y(A, x, y)^2}.
\end{equation}

\begin{table*}[ht]
    \centering
    \small
    \begin{tabular}{lcccccc}
        \toprule
         \multirow{2}{*}{Dataset} & \multicolumn{3}{c}{Faster RCNN}  & \multicolumn{3}{c}{NITF RCNN} \\
        \cmidrule(lr){2-4} \cmidrule(lr){5-7}
        & mAP@0.5 & mAP@0.75 & mAP@[0.5,0.95] & mAP@0.5 & mAP@0.75 & mAP@[0.5,0.95] 
        \\
        \midrule
         KITTI          & 80.87\%  &61.45\%  &55.43\%    & ---- & ---- & ----  \\
        \midrule
        Rainy-KITTI          & 73.17\%  &56.15\%  &49.28\%  
        & 79.16\%($\uparrow$8.1\%)  &59.35\%($\uparrow$5.6\%)  &51.86\%($\uparrow$5.2\%)    \\
        \midrule
        Foggy-KITTI          & 71.88\%  &48.74\%  &42.31\%  
        & 75.01\%($\uparrow$4.4\%)  &51.30\%($\uparrow$5.5\%)  &44.48\%($\uparrow$5.1\%)  \\
        \midrule
        Dark-KITTI          & 62.74\%  &41.39\%  &37.89\% 
        & 65.61\%($\uparrow$4.6\%)  &44.05\%($\uparrow$6.4\%)  &40.07\%($\uparrow$5.7\%)  \\
        \midrule
        Raw-KITTI          & 76.13\%  &58.50\%  &51.76\%  
        & 79.06\%($\uparrow$3.8\%)  &61.91\%($\uparrow$5.8\%)  &54.13\%($\uparrow$4.6\%)  \\
        \midrule
        \bottomrule
    \end{tabular}
    \caption{Evaluation Results of Faster RCNN and NITF RCNN on Different Datasets (The relative percentage improvement on each metric is compared between the NITF RCNN algorithm and Faster RCNN.)}
    \label{tab:map}
\end{table*}

\textbf{Local Gradient Magnitude Difference}

The local gradient magnitude difference between feature maps $A$ and $B$, represented by $/Delta S(x, y)$, is expressed as:

\begin{equation}
\Delta S(x, y) = \left| G(A, x, y) - G(B, x, y) \right|. 
\end{equation}
This metric quantifies the direct structural disparities, highlighting areas where edge and texture information significantly differ due to non-ideal imaging conditions. Essentially, $\Delta S(x, y)$ pinpoints the local discrepancies that the model needs to correct to better align feature maps derived from non-ideal and ideal scenarios.
The term $\exp(-\Delta S(x, y))$ acts as a weighting factor, modulating the contribution of each local discrepancy $\Delta S(x, y)$ to the overall loss. 

When these two terms are multiplied, i.e., $\exp(-\Delta S(x, y))  \cdot  \Delta S(x, y)$,
the exponential decay function in the loss calculation magnifies the impact of smaller discrepancies, directing the model's focus on refining minor but essential structural differences. Simultaneously, it lessens the penalty on larger discrepancies to avoid undue penalization for less critical variances. This mechanism ensures a balanced model training, prioritizing major discrepancies in early phases for overall performance and shifting towards finer adjustments in later stages as feature map discrepancies diminish, facilitating nuanced structural alignment for improved object detection accuracy.

\noindent\textbf{Extended Area Gradient Consistency}

The Extended Area Gradient Consistency term, $\Omega(A, B, x, y, r_{\mathcal{L}})$, scrutinizes the uniformity of gradient transitions within a specified vicinity, thereby assessing broader spatial patterns. It evaluates the consistency of gradient changes across an extended neighborhood, defined by a radius $r_\mathcal{L}$. This radius is adaptively adjusted for each layer in the Faster RCNN feature pyramid, allowing for a multi-scale analysis:

\begin{align}
&\Omega(A, B, x, y, r_{\mathcal{L}}) = \frac{1}{(2r_{\mathcal{L}}+1)^2} \nonumber \\
& \cdot \sum_{i=-r_{\mathcal{L}}}^{r_{\mathcal{L}}} \sum_{j=-r_{\mathcal{L}}}^{r_{\mathcal{L}}} \left| (G(A, x, y) - G(A, x+i, y+j)) \right. \nonumber \\
&\left. - (G(B, x, y) - G(B, x+i, y+j)) \right|.
\end{align}

This extended area gradient consistency ensures that the model not only captures pixel-by-pixel discrepancies, but also appreciates broader spatial patterns and alignments. This multi-scale approach is critical for robust object detection, as it allows the model to recognize and adapt to the variances in object sizes and shapes across different feature map scales.

In summary, EANSDL represents a significant advance in object detection, offering powerful structural insight and correction capabilities.
By skillfully combining evaluations of both immediate and broader spatial contexts, the EANSDL function empowers object detection algorithms that correct feature maps layers through Transfer Learning, such as NITF RCNN, to deliver unparalleled performance. This approach ensures the meticulous alignment and refinement of feature maps, transcending the challenges posed by non-ideal imaging conditions.
This approach not only enhances the model's detection capabilities but also sets a benchmark for feature map analysis and correction in complex visual environments.

%% file: sec/4_experiments.tex
\section{Experiments}
\label{sec:experiments}

\subsection{Dataset Selection and Generation}



In our experimental setup, we combine original and synthesized datasets due to the requirement of having both ideal and non-ideal versions of the same image for transfer learning. This necessity arises from the need to train on ideal images and then adapt to non-ideal conditions. Real non-ideal images, altered to simulate 'ideal' conditions, often lose crucial details due to inherent visual obstructions like rain or fog. Therefore, to maintain content consistency and to ensure a robust training foundation, we opt for real-world images as our ideal dataset and generate synthetic counterparts for the non-ideal scenarios, effectively using high-quality originals to produce less information-dense but contextually aligned images.

The KITTI 2D object detection dataset \cite{geiger2013vision} serves as the foundation for our experiments, known for its real-world driving scenarios, diverse object annotations, and complex urban environments. As the ideal dataset, we utilize KITTI alongside four synthetic datasets—Rainy-KITTI, Foggy-KITTI, Dark-KITTI, and RAW-KITTI—as our non-ideal datasets.

\begin{itemize}
    \item{\textbf{Rainy-KITTI \& Foggy-KITTI:}} For simulating rain and fog conditions, we selected the Rainy-KITTI and Foggy-KITTI datasets \cite{halder2019physics,tremblay2021rain}, recognized for their realistic emulation of these weather effects. The Rainy-KITTI dataset encompasses images under seven distinct rain intensities, ranging from light to heavy downpours. Similarly, the Foggy-KITTI dataset includes images under seven different visibility conditions due to fog. For our experiments, we randomly select an image from each of these conditions to compile our dataset.
    \item{\textbf{Dark-KITTI:}} To generate a dataset simulating low-light conditions, we followed Rashed et al.'s methodology \cite{rashed2019fusemodnet}, utilizing the UNIT \cite{liu2017unsupervised} algorithm for its superior performance in creating realistic night-time images. Using 2000 clear-day images from the KITTI dataset \cite{geiger2013vision} and 2000 night images from the BDD100K dataset \cite{yu2020bdd100k}, we trained a day-to-night model on UNIT and generated the Dark-KITTI dataset.
    \item{\textbf{Raw-KITTI:}} Addressing the challenge of replicating RAW Bayer images, due to the irreversible nature of the Image Signal Processing (ISP) \cite{wei2022vehicle}, we adopted a dataset generation method from \cite{chan2023raw} to create a synthetic color Bayer image dataset, termed Raw-KITTI. This dataset features color channels in the RGB format, ensuring consistency in channel count across all datasets used in our experiments by assigning corresponding colors to each channel of the RAW data.

\end{itemize}

\begin{figure*}[ht]
    \centering
    \includegraphics[width=1\linewidth]{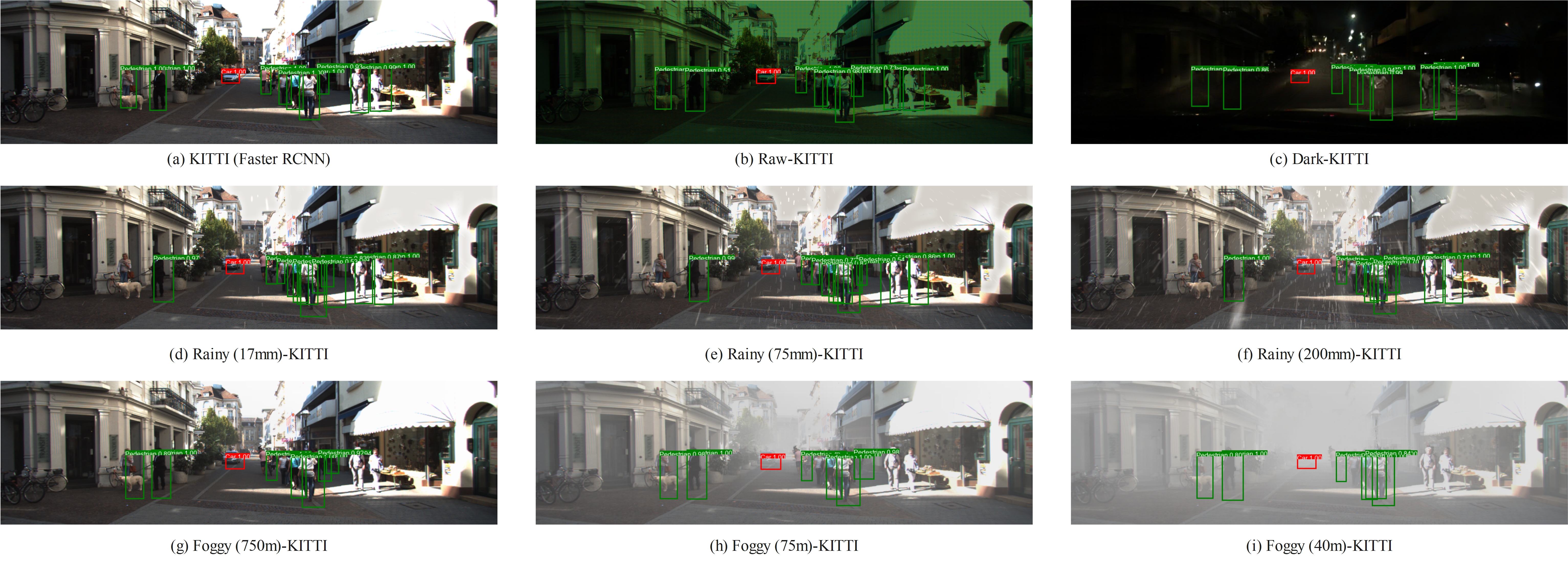}
    \caption{Detection Results of NITF-RCNN on Derivative Images of ID 000332 from the KITTI Dataset, where (a) represents the original image from the KITTI dataset detected using the Faster RCNN algorithm for comparison.}
    \label{fig:detection Results}
\end{figure*}

\subsection{Quantitative Results}

In our experiment, the learning rate was set to 0.005 and the batch size was configured at 8. We allocated 80\% of each dataset for training and reserved 20\% for validation, conducting the training over 100 epochs. The performance evaluation was based on the mean Average Precision (mAP), in accordance with the COCO detection benchmark standards \cite{lin2014microsoft}. Our model's performance was evaluated using three key metrics: mAP@0.5, mAP@0.75, and mAP@[0.5:0.95]. 
The mAP@0.5 and mAP@0.75 metrics represent the mean average precision at Intersection over Union (IoU) thresholds of 0.5 and 0.75, respectively, demanding closer alignment with ground truth for higher values. The mAP@[0.5:0.95] metric, averaging performance across an IoU threshold range from 0.5 to 0.95 with 0.05 increments, provides a comprehensive assessment of model accuracy at various levels of detection precision.

Table \ref{tab:map} encapsulates the evaluation results of Faster RCNN and our NITF-RCNN model across different datasets. It highlights the comparative performance improvements of NITF-RCNN over Faster RCNN under various conditions, showcasing the effectiveness of our FCTL framework.

The results clearly demonstrate the superior performance of NITF-RCNN across all non-ideal imaging conditions, with notable performance gains. Specifically, NITF-RCNN achieved an 8.1\% increase in mAP@0.5, a 5.6\% increase in mAP@0.75, and a 5.2\% improvement in mAP@[0.5:0.95] on the Rainy-KITTI dataset. Similar improvements are observed across Foggy-KITTI, Dark-KITTI, and Raw-KITTI datasets, underlining the model's enhanced detection capabilities in challenging visual scenarios.  

It is noteworthy that across all evaluated datasets, NITF-RCNN exhibits consistent improvements over Faster RCNN in the comprehensive mAP@[0.5:0.95] metric, with relative gains ranging from 4.6\% to 5.7\%. This underscores the effectiveness of the NITF-RCNN model in maintaining high accuracy across various levels of detection precision, especially in non-ideal imaging conditions. Remarkably, on the Raw-KITTI dataset, the mAP@[0.5:0.95] performance of the NITF-RCNN approaches that of the ideal KITTI dataset on the Faster RCNN, with a mere 1.3\% difference. This highlights the significant advancements made by NITF-RCNN in closing the gap between object detection performances in ideal versus non-ideal conditions, showcasing its potential to operate effectively across a broader range of real-world scenarios.

\subsection{Qualitative Results}
Figure \ref{fig:detection Results} displays the detection outcomes on four derivative datasets from the KITTI dataset, specifically for image ID 000332. For the Rainy-KITTI and Foggy-KITTI datasets, we showcase detection results across three different levels of rainfall intensity and varying visibility, respectively.

As a result of integrating the insights derived from Figure \ref{fig:detection Results} and Table \ref{tab:map}, we are able to demonstrate that our methodology yields performance similar to that of Faster RCNN under ideal conditions for the Rainy KITTI and Raw KITTI datasets. In contrast, the performance on the Dark KITTI and Foggy KITTI datasets is relatively inferior. It is hypothesized that the main reason for this discrepancy is that Rainy and Raw KITTI images are more visual discernible than low-light and foggy images, which facilitates easier detection \cite{mai20213d,xiao2020making}.

%% file: sec/5_Conclusion.tex
\section{Conclusion}
\label{sec:conclusion}
This research introduces a pioneering approach in computer vision, particularly in object detection under non-ideal conditions such as low light, adverse weather, or directly from raw Bayer images without ISP. Using the novel concept of FCTL in conjunction with a unique loss function, EANSDL, we demonstrate that the NITF-RCNN model is able to significantly improve the object detection ability in various challenging environments. Our methodology bypasses traditional preprocessing requirements for non-ideal images, directly refining the model's feature maps to closely align with those obtained from pristine RGB datasets. Experimental results demonstrate the efficacy of this approach, which shows a substantial improvement in mAP over conventional methods, thus setting an industry benchmark.

 This study not only strengthens the robustness and accuracy of object detection models under diverse environmental conditions, but also provides new avenues for further research. It is possible for FCTL to be applied outside of autonomous driving, surveillance, and augmented reality, suggesting its potential for other areas where visual data is compromised by conditions that are not ideal. In future research, FCTL may be adapted to address a wider range of imaging challenges, loss function optimization for higher efficiency, and integration of this approach with other frameworks for object detection. The successful application of FCTL heralds a paradigm shift in how visual data is processed, and promises advancements in various applications reliant on accurate and reliable object detection.